\documentclass[sigconf]{acmart}

\usepackage{comment}
\DeclareMathOperator*{\argmin}{arg\,min}

\AtBeginDocument{%
  \providecommand\BibTeX{{%
    \normalfont B\kern-0.5em{\scshape i\kern-0.25em b}\kern-0.8em\TeX}}}

\setcopyright{acmcopyright}
\copyrightyear{2018}
\acmYear{2018}
\acmDOI{10.1145/1122445.1122456}

\acmConference[Woodstock '18]{Woodstock '18: ACM Symposium on Neural
  Gaze Detection}{June 03--05, 2018}{Woodstock, NY}
\acmBooktitle{Woodstock '18: ACM Symposium on Neural Gaze Detection,
  June 03--05, 2018, Woodstock, NY}
\acmPrice{15.00}
\acmISBN{978-1-4503-XXXX-X/18/06}

\begin{document}


\title{Defending against substitute model black box adversarial attacks with the 01 loss}

\author{Yunzhe Xue}
\email{yx277@njit.edu}
\affiliation{%
  \institution{New Jersey Institute of Technology}
  \streetaddress{Department of Computer Science, GITC 4100}
  \city{Newark}
  \state{New Jersey}
  \postcode{07102}
}

\author{Meiyan Xie}
\email{mx42@njit.edu}
\affiliation{%
  \institution{New Jersey Institute of Technology}
  \streetaddress{Department of Computer Science, GITC 4100}
  \city{Newark}
  \state{New Jersey}
  \postcode{07102}
}

\author{Usman Roshan}
\email{usman@njit.edu}
\affiliation{%
  \institution{New Jersey Institute of Technology}
  \streetaddress{Department of Computer Science, GITC 4100}
  \city{Newark}
  \state{New Jersey}
  \postcode{07102}
}


\renewcommand{\shortauthors}{Yang et al.}

\begin{abstract}
Substitute model black box attacks can create adversarial examples for a target model just by accessing its output labels. This poses a major challenge to machine learning models in practice, particularly in security sensitive applications. The 01 loss model is known to be more robust to outliers and noise than convex models that are typically used in practice. Motivated by these properties we present 01 loss linear and 01 loss dual layer neural network models as a defense against transfer based substitute model black box attacks. We compare the accuracy of adversarial examples from substitute model black box attacks targeting our 01 loss models and their convex counterparts for binary classification on popular image benchmarks. Our 01 loss dual layer neural network has an adversarial accuracy of 66.2\%, 58\%, 60.5\%, and 57\% on MNIST, CIFAR10, STL10, and ImageNet respectively whereas the sigmoid activated logistic loss counterpart has accuracies of 63.5\%, 19.3\%, 14.9\%, and 27.6\%. Except for MNIST the convex counterparts have substantially lower adversarial accuracies. We show practical applications of our models to deter traffic sign and facial recognition adversarial attacks. On GTSRB street sign and CelebA facial detection our 01 loss network has 34.6\% and 37.1\% adversarial accuracy respectively whereas the convex logistic counterpart has accuracy 24\% and 1.9\%. Finally we show that our 01 loss network can attain robustness on par with simple convolutional neural networks and much higher than its convex counterpart even when attacked with a convolutional network substitute model. Our work shows that 01 loss models offer a powerful defense against substitute model black box attacks.
\end{abstract}




\maketitle

\section{Introduction}
Adversarial attacks present a  challenge to machine learning algorithms typically based on convex losses. State of the art classifiers like the support vector machine \cite{cortes95} and neural networks \cite{krizhevsky2012imagenet} achieve high accuracies on test data but are also vulnerable to adversarial attacks based on minor perturbations in the data \cite{goodfellow2014explaining,papernot2016limitations,kurakin2016adversarial,carlini2017towards,brendel2017decision}. To counter adversarial attacks many defense methods been proposed with adversarial training being the most popular \cite{szegedy2013intriguing}. This is known to improve model robustness but also tends to lower accuracy on clean test data that has no perturbations \cite{raghunathan2019adversarial,zhang2019theoretically,raghunathan2019adversarial}. 


The robustness of outliers to the 01 loss is well known \cite{bartlett04}. Convex loss functions such as least squares are affected by both correct and incorrectly classified outliers and hinge is affected by incorrectly classified outliers whereas the 01 loss is robust to both \cite{xie2019,icml13optimize}. In addition to being robust to outliers the 01 loss is also robust to noise in the training data \cite{manwani2013noise,ghosh2015making} and under this loss minimizing the empirical risk amounts to minimizing the empirical adversarial risk \cite{lyu2019curriculum,hu2016does} with certain assumptions of noise. We conjecture that these properties may translate to robustness against substitute model black box adversarial attacks that typically succeed in fooling state of the art classifiers \cite{papernot2017practical}.

To test this we first develop stochastic coordinate descent solvers for 01 loss based upon prior work \cite{xie2019}. 
We extend the previous work to a non-linear dual layer 01 loss network that we call MLP01. For the task of binary classification on standard image recognition benchmarks we show that our linear 01 loss solver and the MLP01 network are both as accurate as their convex counterparts, namely the linear support vector machine and the sigmoid activated logistic loss dual layer network. We then subject all methods to a substitute model black box attack \cite{papernot2017practical} and find our 01 loss models to be more robust than their convex counterparts for binary classification on several different image benchmarks. Thus our 01 loss models offer an effective defense against substitute model black box attacks. Below we describe our methods followed by detailed experimental results.


\section{Methods}

\subsection{A dual layer 01 loss neural network}
The problem of determining the hyperplane with minimum number of misclassifications
in a binary classification problem is known to be NP-hard \cite{ben03}.
In mainstream machine learning literature this is called minimizing the 01 loss
\cite{kernel01} given in Objective~\ref{obj1},

\begin{equation}
\frac{1}{2n}\argmin_{w,w_0} \sum_i (1-sign(y_i(w^Tx_i+w_0)))
\label{obj1}
\end{equation}

where $w \in R^d$, $w_0 \in R$ is our hyperplane, and $x_i \in R^d, y_i\in \{+1,-1\}.\forall i=0...n-1$ are our training data. 
We extend the 01 loss to a simple two layer neural network with $k$ hidden nodes and sign activation that we call the MLP01 loss. This objective for binary classification can be given as

\begin{equation}
\small
\frac{1}{2n}\argmin_{W, W_0, w,w_0} \sum_i (1-sign(y_i(w^T(sign(W^Tx_i+W_0))+w_0)))
\label{obj2}
\end{equation}

where $W \in R^{d\times k}$, $W_0 \in R^k$ are the hidden layer parameters, $w\in R^k, w_0\in R$ are the final layer node parameters, $x_i \in R^d, y_i\in \{+1,-1\}.\forall i=0...n-1$ are our training data, and $sign(v\in R^k)=(sign(v_0), sign(v_1),...,sign(v_{k-1}))$. While this is a straightforward model to define optimizing it is a different story altogether. Optimizing even a single node is NP-hard which makes optimizing this network much harder. Note that our weights are real numbers as opposed to binarized neural networks whose weights are constrained to be +1 and -1 or 1 and 0 \cite{galloway2017attacking,courbariaux2016binarized,rastegari2016xnor}. 



\subsection{Stochastic coordinate descent for 01 loss}
We solve both problems with stochastic coordinate descent based upon earlier work \cite{xie2019}. We initialize all parameters to random values from the Normal distribution with mean 0 and variance 1. We then randomly select a subset of the training data (known as a batch) and perform the coordinate descent analog of a single step gradient update in stochastic gradient descent \cite{bottou2010large}. We first describe this for a linear 01 loss classifier which we obtain if we set the number of hidden nodes to zero. In this case the parameters to optimize are the final weight vector $w$ and the threshold $w_0$. 

When the gradient is known we step in its negative direction by a factor of the learning rate: $w=w-\eta\nabla(f)$ where $f$ is the objective. In our case since the gradient does not exist we randomly select $k$ features (set to 64, 128, 256, and 256 for MNIST, CIFAR10, STL10, and ImageNet respectively), modify the corresponding entries in $w$ by the learning rate (set to 0.17) one at a time, and accept the modification that gives the largest decrease in the objective. Key to our search is a heuristic to determine the optimal threshold each time we modify an entry of $w$. In this heuristic we perform a linear search on a subset of the projection $w^Tx_i$ and select $w_0$ that minimizes the objective.

We repeat the above update step on randomly selected batches for a specified number of iterations given by the user. 
We run 1000 iterations to ensure a deep search with an intent to maximize test accuracy. 
In a dual layer network we have to optimize our hidden nodes as well. In each of the 1000 iterations of our search we apply the same coordinate update described above, first to the final output node and then a randomly selected hidden node. In preliminary experiments we find this to be fast and almost as effective as optimizing all hidden nodes and the final node in each iteration. 

Our intuition is that by searching on just the sampled data we avoid local minima and across several iterations we can explore a broad portion of the search space. Throughout iterations we keep track of the best parameters that minimize our objective on the full dataset. The problem with our search described here is that it will return different solutions depending upon the initial starting point. To make it more stable we run it 32 times from different random seeds and use the majority vote for prediction. In the Supplementary Material (\url{http://s000.tinyupload.com/?file_id=07384641501849770229}) we provide full details of our algorithms. 

\subsection{Implementation, experimental platform, and image data}
We implement our 01 loss models in Python and Pytorch \cite{pytorch}, and both MLP and SVM (LinearSVC class) in scikit-learn \cite{scikit}. We optimize MLP with stochastic gradient descent that has a batch size of 200, momentum of 0.9, and learning rate of 0.01 (.001 for ImageNet data). We ran all experiments on Intel Xeon 6142 2.6GHz CPUs and NVIDIA Titan RTX GPU machines (for parallelizing multiple votes). Our SCD01 and MLP01 source codes, supplementary programs, and data are available from \url{https://github.com/zero-one-loss/mlp01}. 


We experiment on four popular image benchmarks: MNIST \cite{lecun1998gradient}, CIFAR10 \cite{krizhevsky2009learning}, STL10  \cite{coates2011analysis}, and ImageNet \cite{ILSVRC15}. Between classes 0 and 1 in MNIST (also digits 0 and 1) we have 12,665 training images and 2115 test images each of size $28\times28$. In CIFAR10 between classes 0 and 1 we have 10,000 $32\times32\times3$ training images and 2000 test ones and in STL10  \cite{coates2011analysis} we have 1000 $96\times96\times3$ training images and 1600 test. In ImageNet classes 0 and 1 contain about 2580 $256\times256\times3$ training images and 100 test ones. We change the split so as to increase the test data size so that we can better train the black box attack substitute model. We divide the training set into two parts: the first containing 1280 for training and 1300 for test. We normalize each image in each benchmark by dividing each pixel value by 255.

\section{Results}
We refer to our linear (no hidden layer) and non-linear (dual layer with 20 hidden nodes) 01 loss models as SCD01 and MLP01 respectively. 
As convex counterparts we select the linear support vector machine (with a cross-validated regularization parameter) denoted as SVM and a dual layer 20 hidden node neural network with sigmoid activation and logistic loss (MLP). 
We use the majority vote of 32 runs for our 01 loss models to improve stability and do the same for SVM and MLP by majority voting on 32 bootstrapped samples. 

We refer to the accuracy on the test data as clean data test accuracy. An incorrectly classified adversarial example is considered a successful attack whereas a correctly classified adversarial is a failed one. Thus when we refer to accuracy of adversarial examples it is the same as $100-attack success rate$. The lower the accuracy the more effective the attack.

\subsection{Substitute model black box attacks}
\begin{figure}[h]
  \centering
  \includegraphics[trim=0 0 0 0, clip, scale=.4]{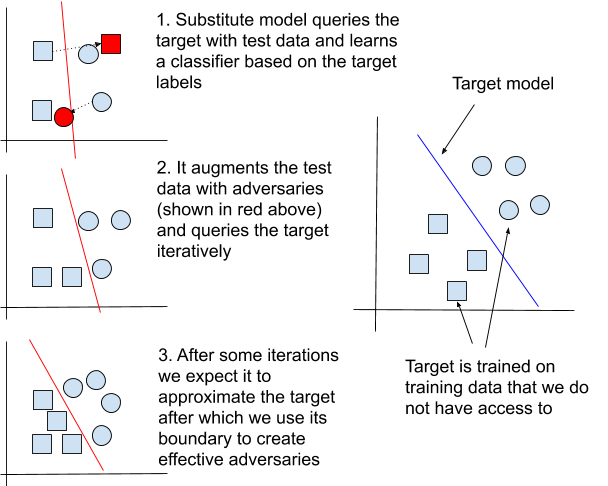} 
  \caption{Substitute model black box attack \label{blackboxattack}}
\end{figure}

We use adversarial data augmentation \cite{papernot2016transferability} to iteratively train a substitute model trained on label outputs from the target model. In each epoch we generate white box adversaries targeting the substitute model with the FGSM method \cite{goodfellow2014explaining} and evaluate them on the target. Since we are considering only binary classification our black box attacks are untargeted. See Supplementary Material (\url{http://s000.tinyupload.com/?file_id=07384641501849770229}) for the full substitute model learning algorithms but it is essentially the method of Papernot et. al. \cite{papernot2016transferability}. Figure~\ref{blackboxattack} gives a visual overview of the substitute model black box attack.
 

\subsubsection{Convex substitute model}

\begin{figure}[!h]
\begin{center}
\centerline{\includegraphics[trim=0 20 0 20, clip, scale=.4]{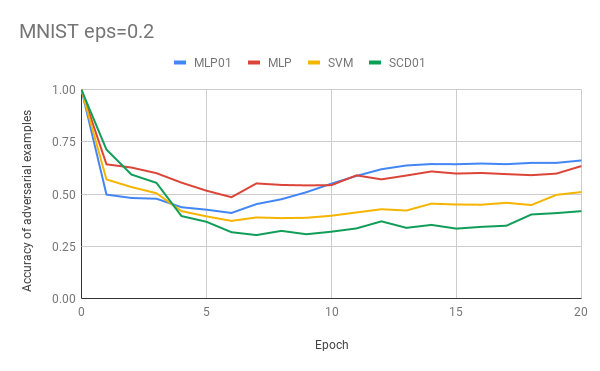}}
\centerline{\includegraphics[trim=0 20 0 20, clip, scale=.4]{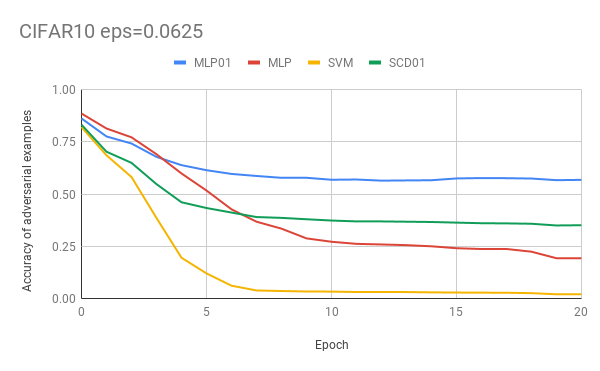}}
\centerline{\includegraphics[trim=0 20 0 20, clip, scale=.4]{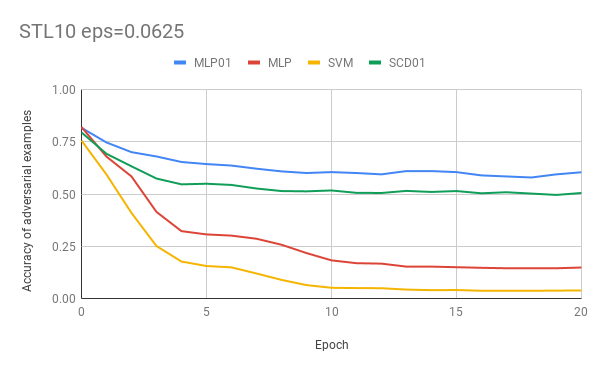}}
\centerline{\includegraphics[trim=0 20 0 20, clip, scale=.4]{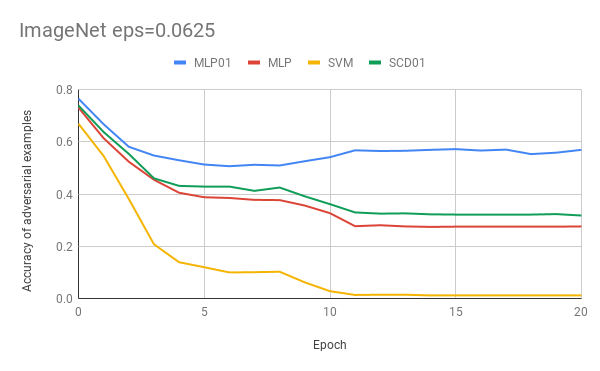}}
\caption{Accuracy of adversarial samples in each epoch of substitute model training. At epoch 0 we have the accuracy of the target model on clean test data.}
\label{blackbox}
\end{center}
\end{figure}

In Figure~\ref{blackbox} we see the accuracy of target models on adversaries generated from a convex substitute model. Specifically we use a three layer neural network as the substitute model with sigmoid activation, logistic loss, and 200 nodes in each hidden layer. We use $\epsilon$ values on MNIST, CIFAR10, STL10, and ImageNet that are typical in the literature. For MNIST $\epsilon=.2$ corresponds to a change of $255\times.2=51$ in each pixel and for CIFAR10, STL10, and ImageNet $\epsilon=.0625$ corresponds to a change of $255\times.0625=16$ in each pixel.

In MNIST (Figure~\ref{blackbox}(a)) we don't see a considerable difference between the 01 loss and convex sibling models although MLP01 has the highest accuracy. On CIFAR10, STL10, and ImageNet we see much more pronounced differences. In CIFAR10 (Figure~\ref{blackbox}(b)) we see that even though both MLP and MLP01 start off with  clean test accuracies of 88\% and 86\% respectively, at the end of the 20th epoch MLP01 has 58\% accuracy on adversarial examples while MLP has 19.3\% accuracy. 

We see similar results on STL10 and ImageNet (Figure~\ref{blackbox}(c) and(d)). Both MLP and MLP01 start off with high similar clean test accuracies but by epoch 20 MLP01 has 60.5\% and 57\% adversarial accuracy on STL10 and ImageNet respectively whereas MLP has 14.9\% and 27.6\%. In the same way the linear 01 loss (SCD01) has similar clean test accuracy to its hinge counterpart in the SVM but is much more adversarially robust on all datasets except for MNIST. 
	

\subsubsection{01 loss substitute model}
\begin{figure}[!h]
  \centering
  \includegraphics[scale=.4]{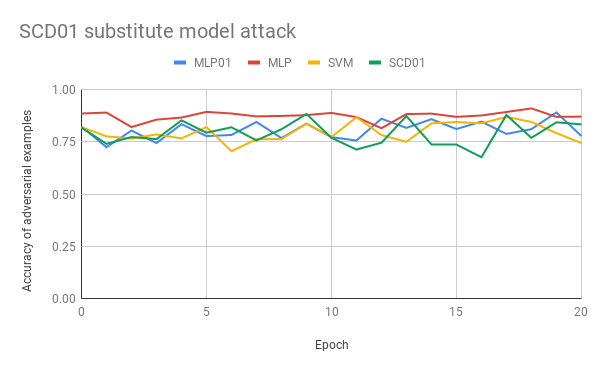}  \\
  \caption{We use SCD01 single run as the substitute model to attack single runs of the target models on CIFAR10 class 0 vs. 1. In epoch 0 are the clean test accuracies. \label{blackbox2}}
\end{figure}

In Figure~\ref{blackbox2} we see the results of a black box attack with SCD01 single run as the substitute model attacking single runs of target models. In each epoch of the substitute model training we use the SCD01 model parameters $w$ to generate adversaries as $x'=x-\epsilon y sign(w)$ where $sign(w)=(sign(w_1),sign(w_2),...,sign(w_{d}))$, $\epsilon$ is the distortion, and $y$ is the label of $x$. This is the same method to generate adversaries from any linear classifier as done previously \cite{papernot2016transferability}. We see that adversaries produced from this model hardly affect any of the target models in any of the epochs. Even when the target is SCD01 and trained with the same initial seed as the substitute the adversaries are ineffective. 

\begin{figure}[h!]
\begin{tabular}{c}
\includegraphics[scale=.4]{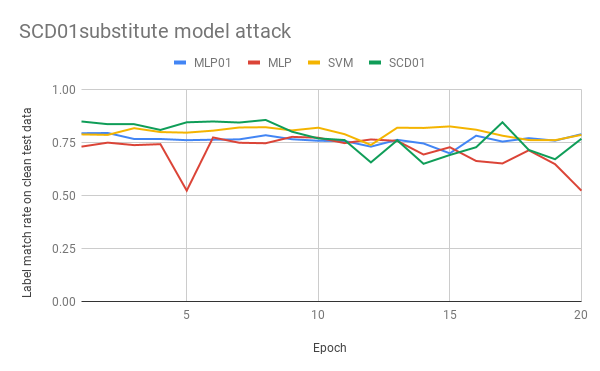}\\
\footnotesize (a) Percent of same labels between substitute and target on clean data \\
\includegraphics[scale=.4]{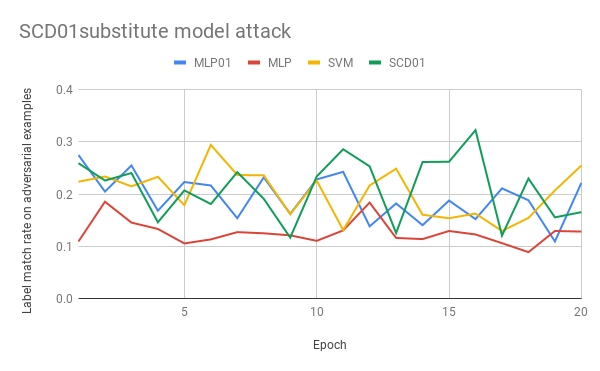}\\
\footnotesize (b) Percent of same labels between substitute and target on adversarial data \\
\end{tabular}
\caption{SCD01 substitute model can approximate the target on CIFAR10 class 0 vs. 1 as shown in the label match rates between them (see (a)). But SCD01 sourced adversaries have a lower match rate which indicates the direction of the SCD01 boundary is  different from the targets and thus its adversaries have  little effect on the targets. \label{01lossattack}}
\end{figure}

Further investigation reveals that the percentage of test data whose labels match between the 01 loss substitute and its target (known as the label match rate) is high but the label match rate on adversarial examples is much lower (see Figure~\ref{01lossattack}). Thus even though the SCD01 manages to approximate the target boundary its direction is different which gives ineffective adversaries. This is due to the non-uniqueness of 01 loss which makes single run solutions different from each other. Thus as a substitute model in black box attacks 01 loss is ineffective even in attacking itself. 

\subsection{Applications: street sign and facial recognition adversarial attacks}
We now turn to two practical problems where adversarial attacks pose a problem. First is the task of street sign detection by autonomous vehicles \cite{sitawarin2018darts} and the second is facial recognition that are used by government and security systems. We consider 2816 train and 900 test $48\times48\times3$ images of street signs of 60 and 120 mph from the GTSRB street sign dataset \cite{Stallkamp-IJCNN-2011} and 1000 train and 1000 test $96\times96\times3$ images of brown and black hair individuals from the CelebA facial recognition benchmark \cite{liu2015faceattributes}. For GTSRB we use a perturbation of $\epsilon=0.03125$ and for CelebA we use $\epsilon=0.0625$ in the black box attack. 

\begin{figure}[h!]
\begin{center}
\centerline{\includegraphics[width=\columnwidth]{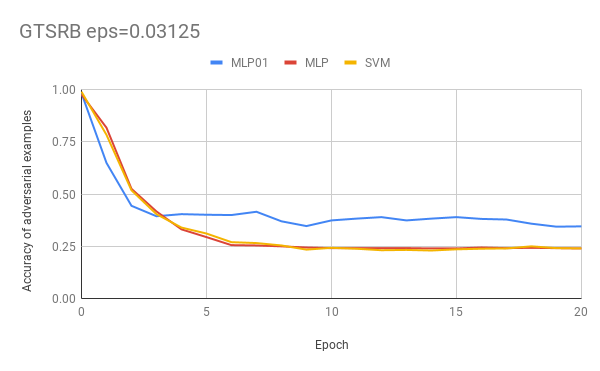}}
\centerline{\includegraphics[width=\columnwidth]{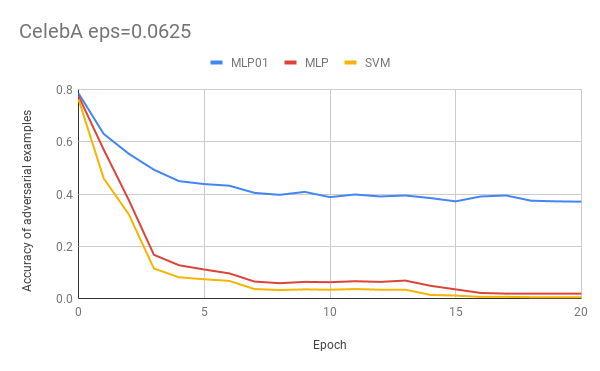}}
\caption{Accuracy of adversarial samples in each epoch of substitute model training. At epoch 0 we have the accuracy of the target model on clean test data.}
\label{applications}
\end{center}
\vskip -0.2in
\end{figure}

\begin{figure}[h!]
\begin{center}
\begin{tabular}{cccc}
Clean & MLP01  & MLP & SVM \\
\includegraphics[scale=.55]{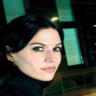} & \includegraphics[scale=.55]{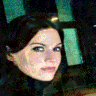} & \includegraphics[scale=.55]{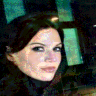} & \includegraphics[scale=.55]{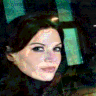} \\
\includegraphics[scale=1.2]{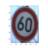} & \includegraphics[scale=1.2]{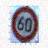} & \includegraphics[scale=1.2]{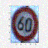} & \includegraphics[scale=1.2]{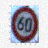} \\
\end{tabular}
\vspace{-.1in}
\caption{Clean and adversarial images from CelebA in the top row and GTSRB in the bottom one. The adversarial images are obtained from the black box attack when targeting MLP01, MLP, and SVM with $\epsilon=0.0625$. The adversarial images shown here are misclassified SVM and MLP but are correctly classified by MLP01.
}
\label{sampleimages2}
\end{center}
\vskip -0.2in
\end{figure}

In Figure~\ref{applications} we see that the MLP01 attains comparable clean test accuracy to SVM and MLP but is more robust just like we saw in the benchmarks above. We show sample adversarial images in Figure~\ref{sampleimages2} obtained from the black box attack with different target models. Even though the adversarial images targeting MLP01 show a visual distortion they are correctly classified by it. 

\setlength{\tabcolsep}{6pt}

\setlength{\tabcolsep}{1pt}
\subsection{Comparison to convolutional neural networks}
We finally compare black box robustness of our 01 loss models to simple convolutional neural networks. We find the pair of CIFAR10 classes where MLP01 achieves the highest test accuracy (this turns out to be classes 6 vs 8). We then train MLP01 and MLP each with 400 hidden nodes and two convolutional neural networks. We take LeNet \cite{lecun1998gradient} and SimpleNet80 that has the convolutional layers from LeNet followed by a single hidden layer of 80 nodes. The latter is a watered down version of LeNet with fewer hidden layers.

We use a convolutional neural network with four convolutional blocks as the substitute model. In each convolutional block we have a $3\times3$ convolutional kernel followed by max pool and batch normalization. In the first, second, third, and fourth layer we have 32, 64, 128, and 256 kernels and a final layer for the output. Thus the substitute model here is much more sophisticated and powerful than the one used in our experiments above.

\begin{figure}[!h]
  \centering
  \begin{tabular}{c}
  \includegraphics[width=\columnwidth]{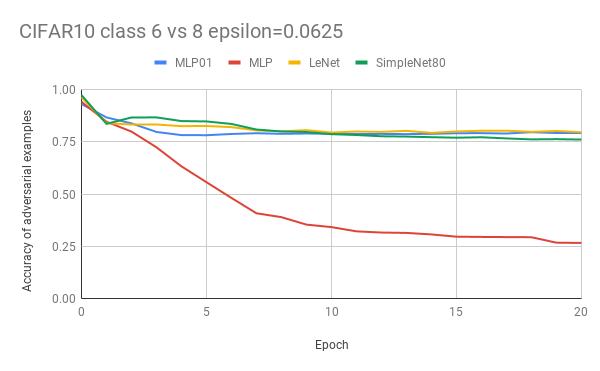} \\
  \includegraphics[width=\columnwidth]{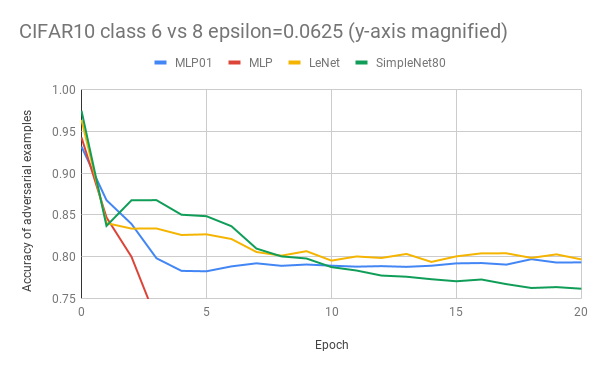} \\
 \end{tabular}
  \caption{Black box attacks with a convolutional neural network of four convolutional blocks as the substitute model. In epoch 0 are the clean test accuracies. We see that MLP01's adversarial accuracy is on-par with convolutional networks but MLP has much lower robustness. \label{cnn}}
\end{figure}

In Figure~\ref{cnn} we see that MLP01 has comparable clean test accuracy to all the convex models including its convex counterpart MLP. However, its robustness is much higher than MLP and on-par with the convolutional models that have the advantage of convolutions over our model. In fact we see that MLP01's robustness even surpasses that of SimpleNet80 after the tenth epoch of the substitute model training. Thus we see that even when the substitute model is a sophisticated convolutional network it is hard to attack our model.

\section{Discussion}
Binarized neural networks \cite{galloway2017attacking,courbariaux2016binarized,rastegari2016xnor} have weights and activations constrained to be near +1 and -1 (or 1 and 0) whereas our model weights are real numbers. The purpose of those networks is efficiency as opposed to robustness. Indeed we see in recent work that binarized networks offer marginal improvements in robustness to substitute model black box robustness on MNIST and none in CIFAR10 (see Tables 4 and 5 in \cite{galloway2017attacking} and Table 8 in \cite{panda2019discretization}). In our case we see large improvements over convex (non-binarized) models on CIFAR10 as well as other datasets. 

There is nothing to indicate that 01 loss models are robust to boundary based black box attacks that do not require substitute model training \cite{brendel2017decision,chen2019hopskipjumpattack}. These attacks are however computationally expensive, require separate computations for each example, and require in the order of thousands of queries per example. In comparison, as we have seen in this study, a transfer based substitute model attack can create effective adversaries within just five to ten epochs (queries) and once it has been trained it can create adversaries without querying the target. Thus it can be more effective and dangerous as an attack method than boundary based ones.

Interestingly the robustness of 01 loss neural network is on-par with simple convolutional models that have the powerful advantage of convolutions which our 01 loss models lack. As future work 01 loss convolutions may be a promising avenue for models with high clean test accuracy and high adversarial accuracy as well. Another promising avenue is to perform adversarial training for our models to further boost robustness \cite{kurakin2016adversarial}.

\section{Conclusion}
Our work here shows that 01 loss models are more robust to substitute model black box attacks than their equivalent convex models on several datasets for binary classification. Thus our models could be used as an effective defense method against transfer based black box attacks.

\bibliographystyle{ACM-Reference-Format}
\bibliography{my_bib}

\end{document}